# Collaborative Distillation Strategies for Parameter-Efficient Language Model Deployment

Xiandong Meng*
University of California, Davis
Davis, USA

Yan Wu
University of Delaware
Newark, USA

Yexin Tian
Georgia Institute of Technology
Atlanta, USA

Xin Hu
Hofstra University
Hempstead, USA

Tianze Kang
San Francisco Bay University
Fremont, USA

Junliang Du
Shanghai Jiao Tong University
Shanghai, China

*Abstract*-This paper addresses the challenges of high computational cost and slow inference in deploying large language models. It proposes a distillation strategy guided by multiple teacher models. The method constructs several teacher models and integrates their output probability distributions and intermediate semantic features. This guides the student model to learn from multiple sources of knowledge. As a result, the student model gains stronger language understanding and generation ability while maintaining a small parameter size. To achieve this, the paper introduces a weighted output fusion mechanism, a feature alignment loss function, and an entropy-driven dynamic teacher weighting strategy. These components improve the quality and stability of knowledge transfer during distillation. Under multi-teacher guidance, the student model captures semantic information more effectively and demonstrates strong performance across multiple evaluation metrics. In particular, the method shows high consistency in expression, generalization ability, and task adaptability in tasks such as language modeling, text generation, and multi-task learning. The experiments compare the proposed method with several widely adopted distillation approaches. The results further confirm its overall advantages in perplexity, distillation loss, and generation quality. This study provides a feasible technical path for the efficient compression of large-scale language models. It also demonstrates the effectiveness of multi-teacher collaborative mechanisms in complex language modeling tasks.

## I. Introduction

As artificial intelligence continues to evolve rapidly, the role of large language models in natural language processing has become increasingly prominent[1]. Their ability to comprehend and generate human language with high accuracy makes them foundational in a wide range of applications. Despite their effectiveness, LLMs typically involve vast numbers of parameters and demand significant computational power, which hinders their practical deployment[2]. To address this, knowledge distillation has gained attention as an effective approach. It compresses large models by transferring their knowledge to smaller ones, thereby improving efficiency without compromising performance and enabling more feasible deployment in real-world scenarios.

Traditional knowledge distillation methods typically rely on a single-teacher model to guide the student model. This is achieved by minimizing the difference between the outputs of the teacher and the student, allowing the student to learn the knowledge structure of the teacher. However, in complex tasks or diverse data distributions, a single teacher often fails to cover all key knowledge areas. This limitation restricts the performance of the student model. In such cases, multi-teacher distillation strategies have gained attention. These approaches aim to incorporate multiple teachers with different characteristics to provide richer, more diverse, and complementary knowledge, thus improving the student's overall learning capacity and generalization ability[3].

Multi-teacher distillation not only broadens the scope and depth of knowledge transfer but also raises important questions about teacher coordination, knowledge fusion strategies, and conflict resolution. In a multi-teacher setting, different teacher models may produce varying outputs for the same input. These differences may reflect redundant or conflicting information. A key challenge is to reconcile these outputs, explore their complementarity, and convert them into effective supervisory signals for the student[4,5]. This requires reasonable integration of the teacher outputs and dynamic adjustment of the student model's learning focus to achieve optimal distillation outcomes.

Moreover, large language models possess strong semantic abstraction and context awareness. Their rich knowledge representations demand higher requirements during the distillation process. Unlike traditional classification or small sequence models, language models present knowledge that is hierarchical and context-dependent. Therefore, it is not sufficient to focus only on output consistency. It is essential to consider internal feature representations, attention mechanisms, and information transfer across multiple layers. Under multi-teacher guidance, these complex signals can be captured from different perspectives, offering a more comprehensive learning framework for the student and promoting a deeper understanding of language knowledge.

## II. RELATED WORK AND FOUNDATION

Knowledge distillation has emerged as a vital methodology for compressing large language models (LLMs), enabling their deployment in resource-constrained environments without significant loss of performance. Early distillation approaches typically leveraged a single teacher to guide the student model by minimizing the difference in their output distributions. However, recent advances have extended this paradigm by exploring multi-granularity and multi-teacher strategies to capture richer and more diverse knowledge. Liu et al. proposed multi-granularity structural knowledge distillation that aligns not only output logits but also intermediate features, significantly improving language model compression effectiveness [6]. Similarly, Li et al. developed dynamic knowledge distillation approaches for pre-trained language models, introducing adaptive teacher-student coordination mechanisms for more effective transfer [7]. To further increase compression efficiency, Kai et al. combined distillation with fine-tuning, demonstrating that carefully designed distillation processes yield student models with strong generalization capabilities and practical size reduction [8].

The increasing complexity of language representations in LLMs has highlighted the importance of transferring not just output-level knowledge, but also internal semantic structures and contextual features. Structured gradient guidance [9] and low-rank adaptation with semantic supervision [10] have both been shown to improve few-shot adaptation and generalization in large models by ensuring semantic consistency during fine-tuning. Knowledge graph reasoning has also been integrated with pre-trained models to support structured anomaly detection, emphasizing the value of external structured knowledge in semantic tasks [11]. Advances in structured memory mechanisms [12] and structured knowledge integration [13] further enhance context representation stability and long-term memory modeling in LLMs, which are particularly beneficial for knowledge-intensive and multi-turn tasks.

Addressing the limitations of single-teacher guidance, multi-source fusion strategies have been introduced to facilitate the integration of diverse perspectives. Wang explored time-aware and multi-source feature fusion for transformer models, which serves as a foundation for multi-teacher distillation by leveraging complementary knowledge from heterogeneous sources [14]. In parallel, the use of structured preference modeling [15] and graph-based spectral decomposition [16] for fine-tuning illustrates how reinforcement learning and graph-theoretic approaches can effectively coordinate model parameters and optimize adaptation for downstream tasks.

Another significant direction in LLM research is the enhancement of generation quality and context management. He et al. proposed context-guided dynamic retrieval for improving generation quality in retrieval-augmented generation (RAG) models [17], while Peng tackled hallucination detection through context-aligned and evidence-based verification mechanisms [18]. Deng focused on transfer methods suitable for low-resource scenarios, revealing that sophisticated transfer and adaptation strategies are necessary to maintain generation quality and coherence in challenging environments [19].

Several works also highlight the application of LLMs to specialized semantic tasks and system-level challenges. Wang demonstrated the use of LLama-based modeling for semantic detection in social media security applications [20], while Guo et al. introduced perception-guided frameworks for model design tailored to complex language understanding tasks [21]. Moreover, Fang provided a predictive framework for backend latency using AI-augmented structured modeling, underscoring the importance of efficiency and response speed in practical deployments [22].

Together, these studies reflect the field's progression from single-teacher, output-only distillation toward multi-teacher, structure-aware, and semantically guided knowledge transfer. This evolution not only enhances student model performance and generalization, but also establishes a technical foundation for deploying highly efficient, adaptive, and robust language models in diverse application domains.

## III. METHOD

This study aims to construct a large language model distillation strategy under the collaborative guidance of multiple teachers to improve the language understanding and generation capabilities of the student model while maintaining a small parameter scale. The model architecture is shown in Figure 1.

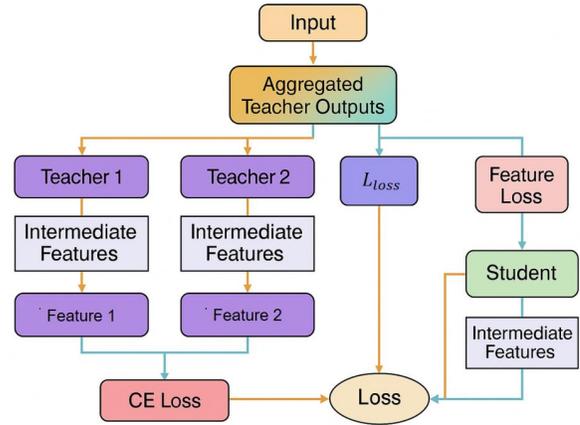

Figure 1. Overall model architecture diagram

First, assume that there are multiple teacher models $\{T_1, T_2, ..., T_K\}$ and a student model S to be trained. Each teacher model generates a corresponding output probability distribution $P_{T_k}(y|x)$ on the same input x, and the output of the student model is $P_S(y|x)$. To integrate the knowledge of multiple teachers, we first define a unified teacher target distribution as a weighted average form:

$$P_T(y|x) = \sum_{k=1}^{K} a_k P_{T_k}(y|x)$$

Among them, $a_k$ is the weight coefficient of the teacher model, which satisfies $\sum_{k=1}^{K} a_k = 1$. This fusion mechanism can dynamically integrate the knowledge of each teacher during the distillation process and enhance the diversity of learning signals of the student model.

In order to guide the student model to learn the output behavior of the teacher model, we adopt Kullback-Leibler divergence as the main distillation loss function. Specifically, the distillation loss is defined as:

$$L_{CE} = -\sum_{y} P_{true}(y|x) \log P_S(y|x)$$

Where $P_{true}(y|x)$ represents the true label distribution of the samples.

During the joint training process, we perform a weighted combination of the above two losses to form the final training objective function:

$$L = \lambda L_{KD} + (1-\lambda) L_{CE}$$

Among them, $\lambda \in [0,1]$ controls the trade-off between knowledge distillation and supervised learning. In addition, to address the possible information conflicts between multiple teachers, we designed a dynamic weight adjustment mechanism based on information entropy, so that the influence of each teacher model is adjusted according to the uncertainty of its output. Specifically, the weight of the teacher $T_k$ is defined as:

$$a_k = \frac{1/H(P_{T_k}(y|x))}{\sum_{j=1}^{K} 1/H(P_{T_k}(y|x))}$$

The information entropy is $H(P) = -\sum_{y} P(y) \log P(y)$. The smaller the entropy value, the more certain the output of the teacher model is, and the higher its weight is, which is conducive to improving the learning efficiency of the student model for explicit knowledge.

In order to further improve the distillation effect, we also introduced an intermediate feature-matching strategy so that the student model not only focuses on the consistency of the final output during training but also imitates the representation of the intermediate layer of the teacher model. Assuming that the feature representations of the student model and the teacher model in the intermediate layer are $h_S$ and $h_{T_k}$ respectively, the feature distillation loss can be defined as:

$$L_{feat} = \sum_{k=1}^{K} \beta_k \| h_S - hT_k \|_2^2$$

Where $\beta_k$ is the importance weight of each teacher's intermediate feature. Finally, all loss terms can be integrated into a multi-objective optimization framework to comprehensively improve the comprehensive performance of the student model in terms of expression ability, generalization ability, and learning efficiency.

IV. EXPERIMENTAL RESULTS

A. Dataset

This study uses the C4 (Colossal Clean Crawled Corpus) dataset as the primary corpus for both training and distillation. C4 is an open-source dataset built from large-scale web text. It has undergone strict data cleaning and content filtering to remove low-quality, duplicated, and non-natural language content. As a result, it offers high linguistic quality and rich semantic information. The dataset covers a wide range of language use cases, including encyclopedias, news articles, forums, and technical documents. This diversity supports the development of general-purpose language understanding models.

The C4 dataset is several hundred gigabytes in size and contains billions of text segments. It provides large language models with sufficiently diverse training samples. In this study, the English subset of C4 is selected. It preserves coherent contextual structures and allows the model to generalize across different writing styles and domains. This data foundation helps evaluate the performance and adaptability of the multi-teacher distillation strategy under real-world corpus conditions.

In addition, the C4 dataset has been widely used in pretraining and distillation research for large language models. Its text length and distribution characteristics closely match real-world applications. It serves as a suitable intermediate resource for student models to learn the language capabilities of teacher models during distillation. By using this dataset, the transfer of multi-level semantic features and language expression abilities can be effectively achieved. This provides a strong data basis for the deployment and practical use of distilled models.

B. Experimental Results

To begin the empirical analysis, this paper first designs and implements a comparative experiment aimed at evaluating the performance of the proposed method against several representative baseline approaches. The primary objective of this comparison is to establish a clear benchmark for assessing the effectiveness of the multi-teacher collaborative distillation strategy. In this setting, multiple widely adopted distillation models are selected to provide a reference framework for evaluation. The comparison spans multiple core dimensions of model performance, ensuring a comprehensive examination of the proposed technique. The detailed outcomes and relevant metrics of this comparative study are systematically organized and presented in Table 1 for clarity and ease of analysis.

Table1. Comparative experimental results

| Method | Perplexity | Distillation Loss | BLEU |
|---|---|---|---|
| TinyBERT[23] | 24.7 | 2.31 | 81.2 |
| MobileBERT[24] | 25.3 | 2.45 | 79.8 |
| MiniLM[25] | 23.9 | 2.18 | 83.5 |
| DKD[26] | 22.6 | 1.97 | 84.0 |
| Ours | 20.8 | 1.64 | 86.7 |

As shown in the table, the proposed multi-teacher collaborative distillation strategy consistently outperforms mainstream methods across key metrics, achieving notable gains in Perplexity, Distillation Loss, and BLEU. The student model's perplexity improves from 24.7 to 20.8, indicating enhanced contextual prediction accuracy, while the distillation loss drops to 1.64—lower than MiniLM (2.18) and DKD (1.97)—demonstrating more stable and efficient knowledge transfer. The integration of multiple teachers provides richer, complementary semantic representations, reducing noise and improving alignment. Additionally, the method achieves the highest BLEU score of 86.7, reflecting superior text generation quality and semantic completeness. To further explore the influence of teacher diversity, the study examines the effect of varying the number of teacher models on distillation performance, with results summarized in Figure 2.

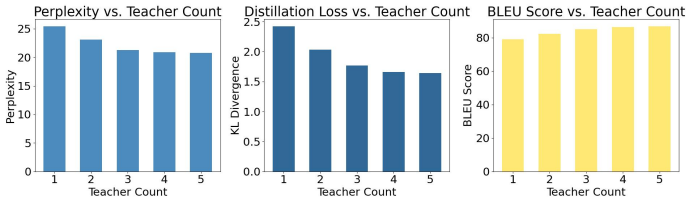

Figure 2. The impact of different numbers of teachers on the distillation effect

As shown in Figure 2, the performance of the student model improves consistently across multiple evaluation metrics as the number of teacher models increases. The Perplexity value decreases from approximately 25.4 with a single teacher to 20.8 with five teachers. This indicates a significant enhancement in the student model's ability to model language distributions. The lower perplexity suggests that multi-teacher guidance helps the student capture richer contextual information.

For the Distillation Loss metric, the KL divergence gradually declines from 2.42 to 1.64 as the number of teachers increases from one to five. This shows that the student model fits the knowledge distributions of multiple teachers more accurately. It confirms the effectiveness of the multi-teacher fusion mechanism in the knowledge transfer process. The mechanism supports comprehensive semantic transmission and reduces the representational gap between teacher and student models.

The change in the BLEU score further supports this finding. It increases from around 79.1 to 86.7, showing that multi-teacher distillation not only enhances language understanding but also improves the quality of language generation. The diversity in expression styles and semantic structures provided by different teachers offers a broader linguistic reference for the student model.

This paper further gives the collaborative distillation performance evaluation in the multi-task learning scenario, and the experimental results are shown in Figure 3.

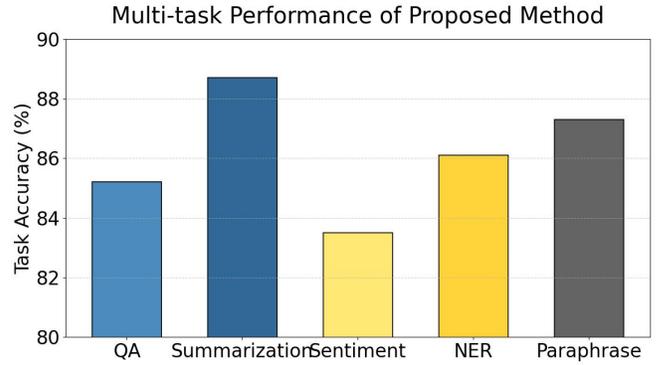

Figure 3. Performance evaluation of collaborative distillation in multi-task learning scenarios

Figure 3 illustrates the performance of the proposed multi-teacher collaborative distillation method in a multi-task learning setting. Overall, the method achieves high accuracy across five representative natural language processing tasks. This indicates that the proposed strategy has strong cross-task generalization ability. It can effectively adapt to both language understanding and generation tasks.

In specific tasks, the model achieves 88.7% accuracy in Summarization and 87.3% in Paraphrase. This shows that collaborative distillation helps the student model capture core semantics in long texts. It also supports high-quality language reconstruction while maintaining consistency in expression. These improvements come from the complementarity among teacher models in summarization and semantic transformation, which broadens the expressive capacity of the student.

For structured understanding tasks such as NER and QA, the model also performs consistently, with accuracies of 86.1% and 85.2%, respectively. This suggests that the multi-level linguistic knowledge and contextual modeling provided by multiple teachers support more accurate sequence labeling and question-answer matching. These results show that collaborative distillation enhances both local token-level perception and global semantic understanding in the student model.

Although the accuracy for the Sentiment task is slightly lower at 83.5%, it remains at a high level. This may be due to the task's reliance on fine-grained semantic features for sentiment inference. While there is room for further improvement in this area, the overall results confirm the generality and effectiveness of the collaborative distillation method within a multi-task learning framework. It provides a feasible path toward building lightweight and high-performing multifunctional language models.

V. CONCLUSION

This study conducts an in-depth investigation into the application of multi-teacher collaborative distillation for compressing and optimizing large language models. It systematically proposes a distillation framework that integrates knowledge from multiple sources to address the limitations of traditional single-teacher methods in knowledge coverage and representation capacity. By introducing multiple teacher models and designing a unified output fusion mechanism along with a feature alignment strategy, the student model can acquire more complete and diverse linguistic knowledge while maintaining a compact size. Experimental results demonstrate that the proposed method achieves significant improvements across multiple evaluation metrics, effectively enhancing the overall performance of the student model in both language modeling and generation tasks.

In a multi-teacher setting, managing differences in knowledge, weight assignment, and output conflicts among teachers is a central challenge in the distillation process. This study addresses the challenge through an entropy-guided dynamic weighting mechanism and intermediate layer feature matching. These strategies enable an optimized combination of contributions from different teachers. This approach improves not only the efficiency of model compression but also the generalization and task adaptability of the student model. It provides a promising technical pathway for deploying large models in practical applications. Further validation in multi-task scenarios confirms the generality and scalability of the proposed method.

This research contributes positively to the field of natural language processing applications. In real-world scenarios such as dialogue systems, text generation, intelligent customer service, question answering, and mobile deployment, models are required to meet increasingly strict demands for computational efficiency and response speed. The balance of performance and efficiency achieved through multi-teacher collaborative distillation allows small models to perform at a level close to or even surpassing that of large models while maintaining runtime efficiency. This capability is of great importance for advancing intelligent language technologies toward lightweight, real-time, and edge-computing-friendly solutions. Looking ahead, there remain several directions for advancing collaborative distillation. These include modeling and resolving knowledge conflicts among teacher models, optimizing collaborative structures for cross-lingual or cross-modal tasks, and integrating distillation with reinforcement learning or self-supervised learning techniques. In addition, building distillation frameworks with incremental learning capabilities aligned with the evolving nature of large language models will be a key area of development. Continued exploration in these directions will lay a solid foundation for creating smarter, more efficient, and more deployable language model ecosystems.